# Classification of Microplastic Particles in Water using Polarized Light Scattering and Machine Learning Methods

## Authors

Leonard Saur[1]*, Marc von Pawlowski[1,2,] Ulrich Gengenbach[1], Ingo Sieber[1], Hossein Shirali[1], Lorenz Wührl[1], Xiangyu Weng[3], Rainer Kiko[3], Christian Pylatiuk[1]*

1) Institute for Automation and Applied Informatics (IAI), Karlsruhe Institute of Technology (KIT), Hermann-von-Helmholtz-Platz 1, 76344 Eggenstein-Leopoldshafen, Germany
2) TÜV Süd Product Service GmbH, Ridlerstraße 65, 80339 München, Germany
3) GEOMAR Helmholtz Centre for Ocean Research Kiel, Wischhofstraße 1-3, 24148 Kiel, Germany

* Corresponding authors: leonard.saur@kit.edu, pylatiuk@kit.edu

# Abstract

The detection and classification of microplastics in water remain a significant challenge due to their diverse properties and the limitations of traditional optical methods. Standard spectroscopic techniques often suffer from the strong infrared absorption of water, while many emerging optical approaches rely on transmission geometries that require sample transparency. This study presents a systematic classification framework utilizing 120° backscattering reflection polarimetry and deep learning to identify common polymers (HDPE, LDPE, and PP) directly in water. This backscattering-based approach is specifically designed to analyze opaque, irregularly shaped particles that lack distinguishable surface features under standard illumination. To ensure high-fidelity data, we introduce a feedback review loop to identify and remove outliers, which significantly stabilizes model training and improves generalization. This framework is validated on a dataset of 600 individually imaged microplastic fragments spanning three polymer types. Our results evaluate the distinct contributions of the Angle of Linear Polarization and the Degree of Linear Polarization to the classification process. By implementing a late fusion architecture to combine these signals, we achieve an average test accuracy of 83%. Finally, a systematic feature hierarchy analysis reveals that the convolutional neural network relies on internal polarization textures associated with the particles' microstructure, rather than on macro-contours, with classification accuracy declining by over 40% when internal structure is removed. This demonstrates that the system extracts polarization-dependent internal structural information that is inaccessible to conventional intensity-only imaging methods.

# 1. Introduction

Microplastics (MPs) are typically defined as plastic particles smaller than 5 mm, with lower size limits ranging from 1 µm to 100 nm depending on the definition (Enfrin et al., 2021). They are recognized as one of the most pervasive classes of micro-pollutants and have been evidenced in marine, freshwater, and drinking water systems worldwide (Avio et al., 2017; Pivokonsky et al., 2018). MPs arise from two main sources: Primary microplastics that are intentionally manufactured, such as beads and fibers used in cosmetics and textile industries, and secondary microplastics formed by fragmentation of larger plastic materials through weathering factors including UV-light, fluctuating temperatures, biofouling, hydrolysis, and mechanical stress (Jahnke et al., 2017).

Microplastics are ingested by a wide variety of aquatic organisms, raising concerns about trophic transfer and human exposure. Documented biological impacts include oxidative stress, inflammation, impaired organ function, and toxic responses to living cells, although the extent of risks to human health remains under debate (Rahman et al., 2021; Sharma and Chatterjee, 2017; Winiarska et al., 2024).

Beyond environmental ingestion, liquid media are widely used in medicine for drug application and resuscitation, where the presence of contaminants poses a direct systemic threat. Recent evidence has already identified and quantified microplastic particles within medical infusions, illustrating a clear pathway for human exposure (Liu et al., 2024). These findings highlight an urgent need for expanded monitoring of such clinical sources, as existing analytical gaps require methods capable of characterizing particles without compromising the probe or the integrity of the sample (Von Pawlowski et al., 2026). These concerns highlight the need for reliable methods to monitor MPs directly in water.

The ability to monitor microplastics directly inside a water matrix, without the need for prior filtration, drying, or chemical digestion, is essential for high-fidelity environmental assessment. Traditional ex-situ analysis workflows are not only labor-intensive but also introduce significant risks of secondary contamination and sample loss during the multi-step preparation process (Brander et al., 2020). Furthermore, removing particles from their aqueous state can alter their surface properties or cause fragile aggregates to fragment, leading to results that do not accurately reflect the real-world distribution of pollutants. Developing methods that can operate in-situ allows for continuous monitoring and the capture of dynamic pollution events that are often missed by discrete grab-sampling techniques.

The identification of microplastics in-situ is governed by a fundamental trade-off between chemical specificity and operational robustness. While existing optical and spectroscopic methods provide valuable data, they are often limited by the physical properties of the water matrix or the particles themselves.

This categorization reflects well-documented physical limitations across established analytical frameworks. Destructive analytical methods such as pyrolysis–gas chromatography–mass spectrometry (Py-GC-MS) provide accurate chemical identification but are fundamentally incompatible with in-situ monitoring as they require complex sample extraction. While precise in measuring elemental mass concentrations, they are inherently destructive and labor-intensive, preventing real-time monitoring (Fischer and Scholz-Böttcher, 2019; Rauert et al., 2025)

Spectroscopic methods like Fourier-transform infrared (FTIR) spectroscopy are hindered by the strong dipole moment of the water molecule, which results in intense infrared absorption bands that obscure the polymer signal (Al Alwan et al., 2024; Shim et al., 2017). Beyond spectral masking, IR-based analysis of aqueous suspensions is further constrained by particle concentration. Standard infrared spectroscopy is frequently precluded in aqueous environments, as the dominant absorption bands of the water matrix effectively mask the spectral signatures of microplastics present at low environmental concentrations.

Raman spectroscopy avoids water absorption but is limited by an inherently small scattering cross-section, which necessitates high-intensity laser excitation that can trigger sample degradation or be obscured by the stochastic fluorescence of organic co-contaminants (Araujo et al., 2018; Kniggendorf et al., 2019; Schymanski et al., 2021).

Emerging optical techniques similarly face geometry-dependent constraints. Digital holography and transmission-mode polarimetry provide high-resolution morphological and phase data, yet they are constrained by their reliance on a ballistic light path, which fails to characterize opaque or heavily weathered particles that diffuse or extinguish incident light rather than transmitting it through the material (Mutuku et al., 2024; Zhu et al., 2024). In the case of holography, extreme sensitivity to phase noise and mechanical vibrations often requires a shielded flow or bypass system to ensure environmental stability during acquisition. Furthermore, exclusively transmission-based polarization often relies on birefringence measurements; however, studies indicate that such signals are frequently artifacts of internal manufacturing stress and vary significantly with particle orientation, rather than being a stable indicator of material chemistry (Matsumoto and Bogue, 1977; Sokkar et al., 2013)

To address these limitations, this study presents a reflection-based polarimetric imaging system designed for the in-situ identification of microplastics. Our approach departs from these constraints by shifting the sensing geometry from transmission to a 120° reflection configuration. While bulk scattering methods have demonstrated that reflected polarized light intensity correlates directly with particle type and concentration, they provide only an integrated intensity signal where the complex interference from varying particle geometries makes it impossible to distinguish single microplastics from natural suspended solids (Liu et al., 2021). The principle of using polarimetric imaging to resolve structural information beyond the reach of standard intensity imaging is well-established in medical diagnostics, particularly in the non-invasive detection of skin cancer. In that field, polarized light is utilized to differentiate between malignant and healthy tissues that appear visually identical by resolving differences in their sub-surface scattering properties (Jacques et al., 2002; Tchvialeva et al., 2012). By employing a reflection-based polarization imaging framework, we advance this methodology by resolving the actual spatial structure of individual particles, extracting high-dimensional polarimetric information that remains obscured in bulk measurement systems. To address these needs, the present study explores a 120° reflection-based polarization imaging framework designed to characterize single microplastic particles directly in water. The implementation of a 120° scattering angle is supported by previous studies demonstrating that backscattered polarimetric signals at this geometry are highly sensitive to refractive index variations and surface textures (Koestner et al., 2024; Liu et al., 2021).

The choice of Angle of Linear Polarization (AOLP) and Degree of Linear Polarization (DOLP) as primary features is supported by their ability to encode material-specific surface characteristics and orientation even in complex environments (Kupinski et al., 2019). As demonstrated in other polarimetric domains, these signatures can distinguish between materials that appear visually identical in standard intensity imaging (Hu et al., 2016).

Rather than estimating composition from refractive indices (Twardowski et al., 2001), our method utilizes a Convolutional Neural Network (CNN) to resolve these material-specific textures. The use of Division of Focal Plane (DoFP) cameras facilitates a snapshot-based acquisition where the primary throughput limitation is restricted only by the hardware framerate. This directly contrasts with the prolonged exposure times inherent to Raman spectroscopy, where typically only ~1% of incident intensity contributes to the inelastic scattering signal (Kniggendorf et al., 2019; Schymanski et al., 2021). By integrating this snapshot polarimetric approach with deep learning classification, we provide a robust framework for the real-time identification of chemically similar polymers in dynamic aquatic environments. The technical positioning of this work in relation to these established methods is summarized in Table 1.

*Table 1: Comparative analysis of microplastic identification methods and their suitability for in-situ water analysis.*

| Method | Type of Data | Primary Limitations in Water | Measurement Constraints | Key References |
|---|---|---|---|---|
| **Pyrolysis Gas Chromatography–Mass Spectrometry (Py-GC/MS)** | • Elemental mass concentration<br>• Bulk compositional data | • Cannot be performed in-situ<br>• Requires total water removal and complex sample preparation<br>• Requires extensive pre-cleaning. | • Destructive: Mass-based data only, not particle counts<br>• One measurement per run<br>• Cannot be combined with imaging | (Fischer and Scholz-Böttcher, 2019; Rauert et al., 2025) |
| **Fourier-Transform Infrared Spectroscopy (FTIR)** | • Particle count and chemical type per particle<br>• Chemical imaging maps (μFTIR) | • Water Absorption: Strong mid-IR absorption by water masks polymer signals.<br>• Biofilms: Organic coatings on wet particles cause spectral interference. | • Time-Intensive: Hours to days for full-filter scanning.<br>• Sampling: Attenuated Total Reflection (ATR) requires dry, physical contact. | (Al Alwan et al., 2024; Shim et al., 2017; Veerasingam et al., 2021) |
| **Raman Spectroscopy** | • Particle count + polymer type per particle<br>• Reveals additives and pigments | • Severe fluorescence interference from organic matter<br>• Humic substances, and dyes | • Long scan times for area mapping<br>• Requires fluorescence quenching<br>• Laser may degrade some polymers | (Jones et al., 2019; Kniggendorf et al., 2019; Schymanski et al., 2021) |
| **Polarized Bulk Backscattering** | • Aggregate optical scattering signal<br>• Semi-quantitative concentration estimate | • Natural suspended sediment, bubbles, and biological particles produce indistinguishable signals<br>• Calibration is matrix-dependent | • Only aggregate concentration proxy<br>• No particle-level, chemical, or morphological data; | (Kumar et al., 2026; Liu et al., 2021) |
| **Digital Holography** | • Particle count<br>• 3D morphology (size, shape, refractive index via phase-contrast maps) | • Material classification based on transmission birefringence<br>• Turbid water degrades hologram quality | • Computationally demanding reconstruction<br>• Field-portable systems emerging but not standardized; ML classifiers needed | (Huang et al., 2023; Li et al., 2024; Montandon and Nicolls, 2024; Zhu et al., 2024) |
| **Transmission Polarimetry** | • Polarization state changes (retardance, depolarization)<br>• Birefringence-based classification | • Requires optical access in transmission<br>• Opaque particles cannot be measured<br>• Suspended sediment creates depolarization noise | • Signal depends on particle thickness and crystallinity, not just polymer type<br>• Calibration libraries less developed than spectral databases | (Valentino et al., 2022; Zeng et al., 2025) |
| **Reflection Polarimetry** | **• Spatially resolved polarization state changes in reflected/scattered light** | **• Requires detectable reflection from particle surfaces** | **• Non-direct measurement**<br>**• Requires labeled image datasets for classification** | **This work** |

# 2. Methods

## 2.1. Microplastic sample preparation

To evaluate the classification framework, non-additivated industrial polymer granules of Polypropylene (PP), High-Density Polyethylene (HDPE), and Low-Density Polyethylene (LDPE) were sourced from LyondellBasell. To simulate realistic microplastic geometries characteristic of environmental surface wear, non-additivated industrial PP, HDPE, and LDPE granules underwent mechanical fragmentation using an abrasive technique under continuous aqueous cooling. This approach produces abraded morphology analogous to Taber abrader and custom abrasion protocols that avoid cryomilling-induced artifacts, while preventing thermal degradation of the polymer structure (Sipe et al., 2024). The resulting fragments were sieved to isolate a size fraction between 50 µm and 300 µm. This size range was selected because particles below 300 µm represent the vast majority of microplastics by count in aquatic environments (Koelmans et al., 2019), while the 50 µm lower limit was chosen to focus on a consistent size group for the current optical setup, though smaller particles could be addressed in future using different magnification objectives.

All particles used in this study are irregularly shaped, colorless, and opaque, lacking distinctive surface, size or color features suitable for visual classification (Figure 1). The prepared microplastic fragments were stored in purified water for four weeks prior to imaging to minimize aggregation and sedimentation.

Even after prolonged immersion, small air bubbles are frequently observed adhering to particle surfaces, indicating that gas entrapment at microplastic-water interfaces is a persistent phenomenon. These bubbles are not removed prior to imaging, as they represent realistic challenges encountered when handling hydrophobic microplastic particles in aqueous environments (Joshua et al., 2025). In total, 600 particles are imaged and analyzed. For each of the three materials (PP, HDPE, and LDPE) 200 particles were used. As illustrated in Figure 1 standard digital microscopy (Keyence VHX 7000) provides insufficient detail for material identification, as the images reflect only superficial morphological traits that are not enough to classify the material (Lusher et al., 2020). This lack of clear visual markers confirms that basic optical inspection cannot reliably distinguish between polymer types, as many particles appear visually identical despite differing chemical compositions. Consequently, this study employs polarization-sensitive imaging to resolve structural data that remains hidden in standard microscopy, providing the necessary input for the classification framework.

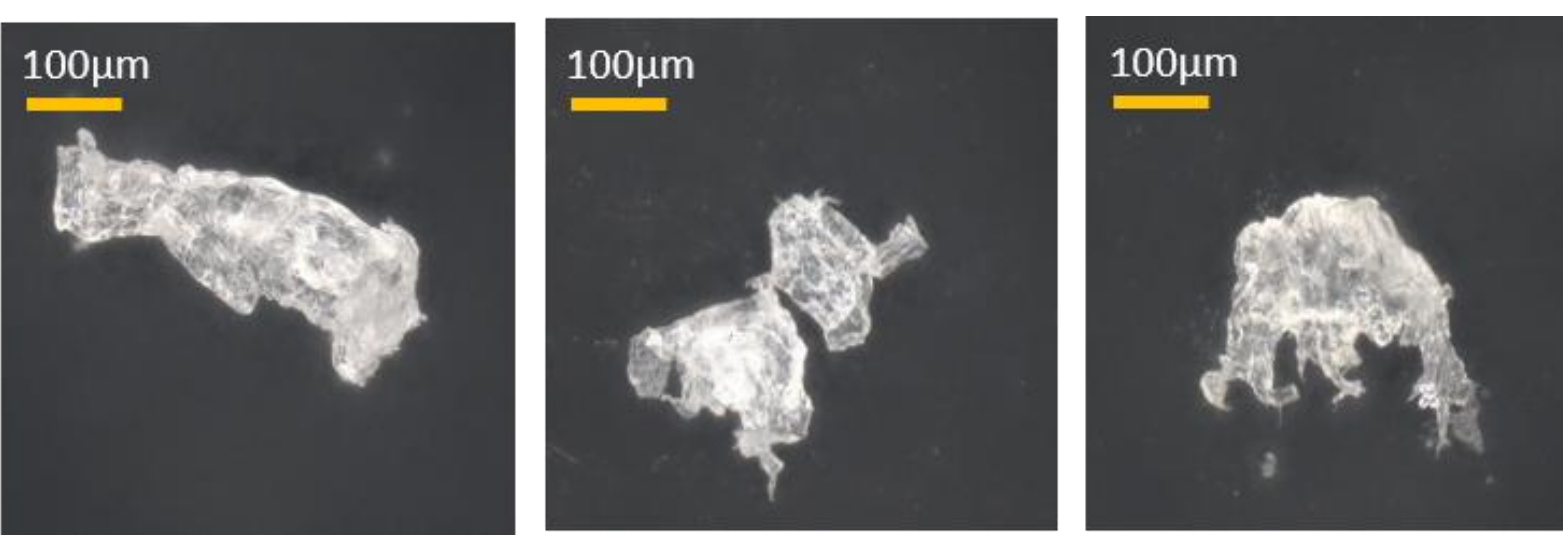

*Figure 1: Example images of classified microplastic particles from left to right: PP, HDPE, and LDPE imaged under a Keyence digital microscope with unpolarized ring light.*

## 2.2. Integrated Analytical Pipeline

The methodology of this study is structured as a multi-stage pipeline designed to transition from physical particle image capture to refined polymer classification. As illustrated in Figure 2 the workflow encompasses four primary phases: **Image acquisition**, utilizing reflection-based polarimetric imaging, **preprocessing and signal reconstruction**, where raw sensor data is converted into Stokes parameters; the **feedback review loop (FRL)**, an outlier-driven refinement process to ensure dataset integrity; and the final evaluation of the classification results using a **feature robustness test**, where the final model's robustness is tested against systematic signal degradation. The following sections detail the technical implementation of each stage.

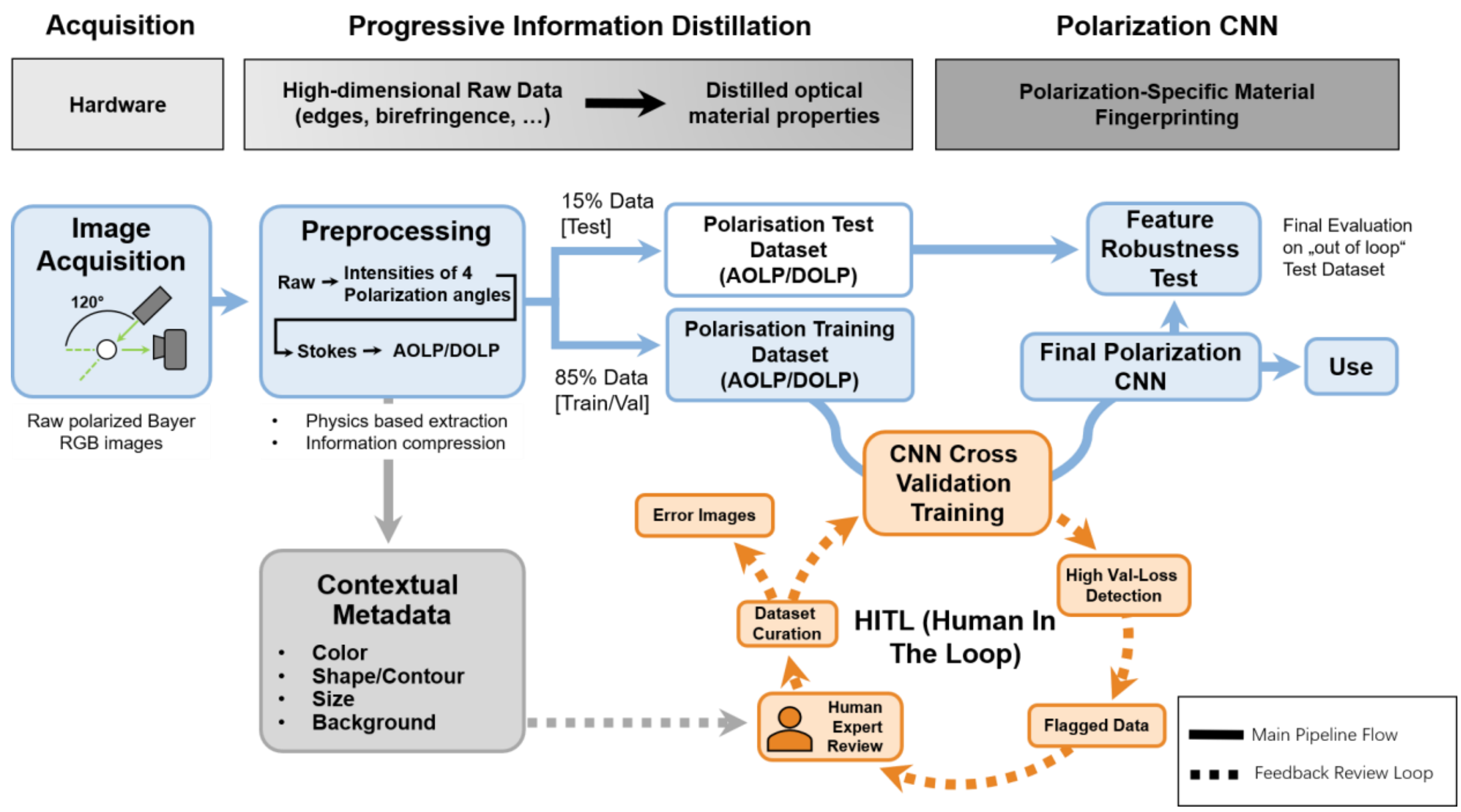

*Figure 2: Schematic of the progressive information distillation and Polarization CNN pipeline. The workflow initiates with multi-angle image acquisition and Stokes-based preprocessing. A distinctive Human-In-The-Loop (HITL) feedback cycle facilitates dataset curation by flagging high-loss samples for expert review, ultimately refining the data foundation for the final CNN classification and feature robustness analysis.*

## 2.3. Optical Setup and Image Acquisition

The experimental configuration is shown in Figure 3. Illumination is provided by a linearly polarized 532 nm diode laser operated at 1 mW. To establish a consistent reference for the polarimetric measurements, the laser is fixed at a 0° polarization plane, providing vertical polarization perpendicular to the sample plane. To ensure uniform interaction with the microplastic fragments, beam focusing optics were utilized to maintain a spot size of approximately 500 µm in the sample plane (see Supplementary Information A for details on laser stability and beam divergence). Scattered light from the particles is collected at 120° relative to the incident beam, a geometry chosen to enhance material-dependent polarization contrast while reducing particle size scattering effects (Boss and Pegau, 2001; Twardowski et al., 2001).

Detection optics are mounted on an adjustable rail system to precisely control angle and working distance. Imaging is performed with a Krüss MSZ5000-T-S stereo zoom microscope equipped with 10× wide-field eyepieces, a 0.7–4.5× zoom objective, and a 2× auxiliary lens, yielding a magnification range of 14× to 90×. At maximum zoom, the field of view corresponds to approximately 801 µm with an effective pixel resolution of 0.39 µm/pixel. A long-working-distance objective ensures clearance for the sample chamber while maintaining high-resolution focusing.

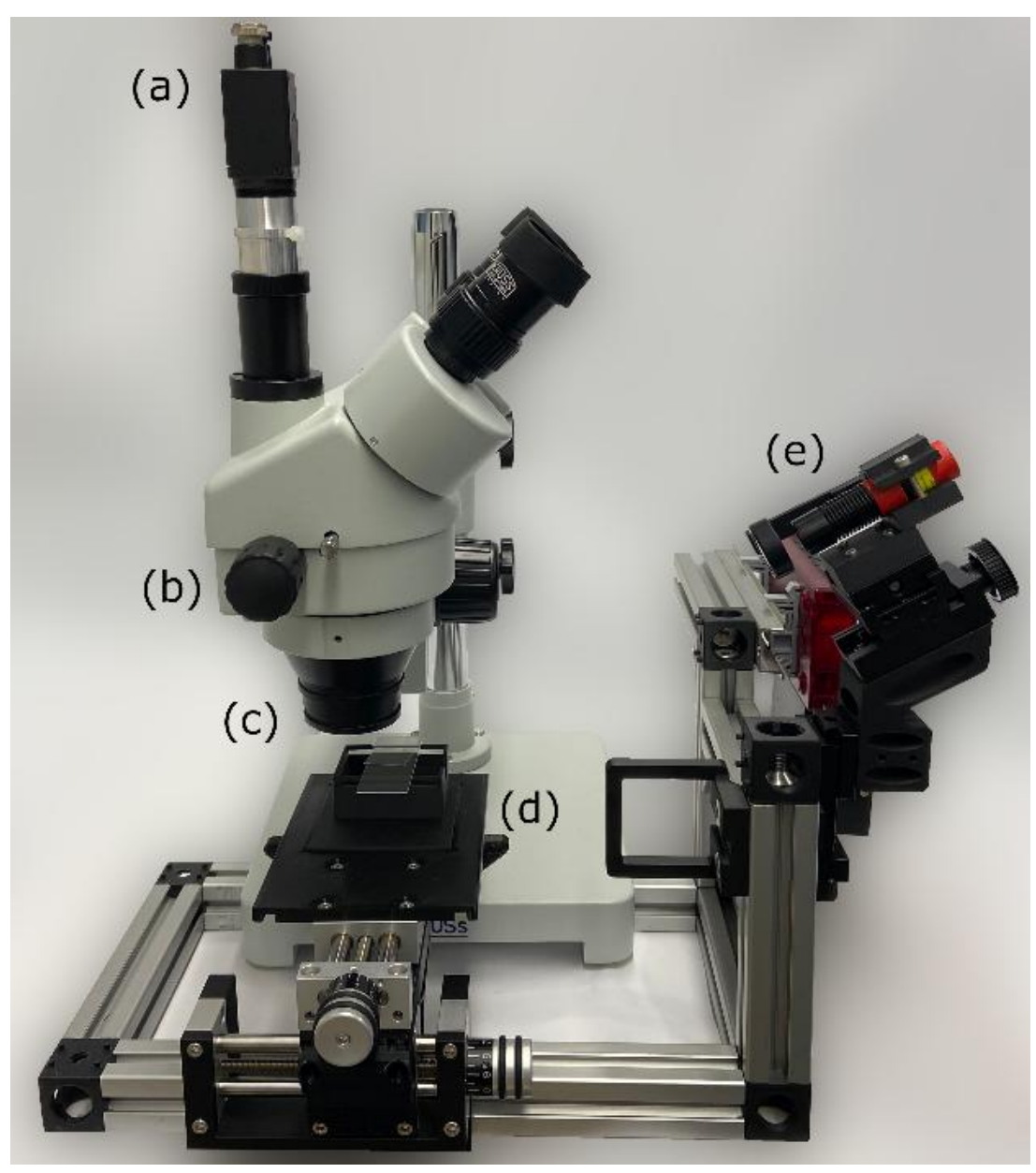

*Figure 3: (a) Polarization camera, (b) adjustable magnification, (c) 2x objective, (d) precision stage and (e) Laser with focusing optics*

A drop of the sample suspension is placed between, and enclosed by, two stacked microscope slides. The samples for analysis are fully immersed in water, with a reference distance of approximately 300 µm derived from the largest particles on a single slide. Diffuse background illumination is provided by white light-emitting diode (LED) strips beneath a diffuser plate positioned 20 mm from the LEDs, ensuring homogeneous illumination across the sample plane.

Image acquisition is performed using the LUCID Vision Labs Triton™ TRI050S-QC polarization camera, which incorporates a Sony IMX250MYR CMOS sensor utilizing Sony's Polarsens Division-of-Focal-Plane (DoFP) technology. Raw polarization intensities at 0°, 45°, 90°, and 135° are captured with the LUCID Arena SDK (Software Development Kit). All automatic functions (auto-exposure, auto-gain, image enhancement) are disabled, and fixed exposure, gain, and white balance settings are used throughout all experiments. To maintain consistent image sharpness across the lateral extent of the sample area, minor axial stage adjustments were performed for each particle to compensate for micro-scale variations in slide planarity. Ambient temperature during measurements is 22–27°C, and the setup is partially enclosed to suppress stray light. Because the excitation wavelength is 532 nm, only the green channel of the camera's bayer array is extracted for analysis, maximizing signal-to-noise ratio.

## 2.4. Preprocessing and Signal Reconstruction

The native output of the green color channel of the DoFP polarization sensor consists of a pixel-level mosaic, where each pixel records intensity at only one of the four analyzer orientations. To reconstruct four full-resolution polarization images required for Stokes analysis, polarization demosaicing is performed. We applied Fourier-Domain Zero-Padding (FDZP) interpolation, a mathematically rigorous approach that aims to approximate the ideal continuous signal from discrete samples (Mihoubi et al., 2018; Qiu et al., 2021). This method preserves the integrity of polarization signals across the DoFP mosaic and provides a robust baseline for subsequent Stokes parameter estimation. Its known trade-off is the potential for minor ringing artifacts near sharp edges. A detailed description of the interpolation algorithm is provided in Supplementary Information B.

The image processing workflow is summarized in Figure 4. From the reconstructed polarization images ($I^{0}, I^{45}, I^{90}, I^{135}$), the linear Stokes parameters are computed as (Collett, 2005):

$$S_0 = I^0 + I^{90},\ S_1 = I^0 - I^{90},\ S_2 = I^{45} - I^{135}$$

From these, the two key polarization descriptors are derived:

$$\mathrm{DOLP} = \frac{\sqrt{\left({S_1}^2 + {S_2}^2\right)}}{S_0} \qquad (Eq{:}\,1-1)$$

$$\mathrm{AOLP} = \frac{1}{2} \times \arctan\left(\frac{\mathrm{S}^2}{\mathrm{S}^1}\right) \qquad (Eq{:}\,1-2)$$

Only pixels with a nonzero value for $S_0$ and $S_1$ are evaluated, ensuring that DOLP and AOLP are computed exclusively in regions with measurable signal, and no division by zero error occurs. These descriptors capture polarization contrast independent of brightness or color and serve as the feature basis for classification (Collett, 2005).

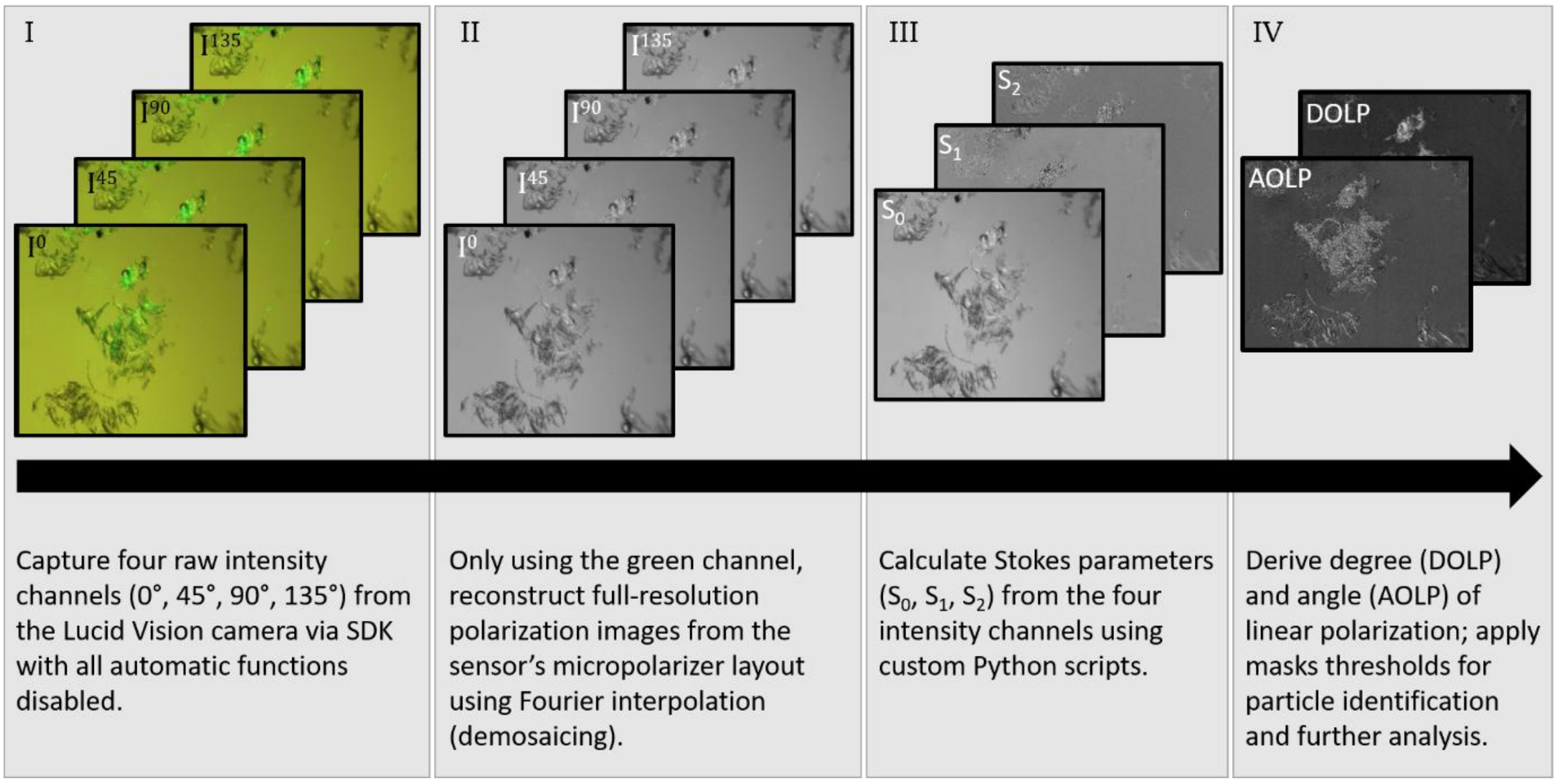

*Figure 4: Workflow for microplastic polarimetric analysis. (I) Raw intensity capture; (II) Fourier de-mosaicing; (III) Stokes parameter calculation; (IV) Generation of DOLP/AOLP maps and particle mask application.*

To prepare the data for machine learning, individual particles are segmented from the background using an automated image-processing pipeline (see Supplementary Information C). The workflow comprises grayscale conversion, denoising, local contrast enhancement, edge detection, and morphological refinement, resulting in filled binary masks of isolated particles. These masks are applied to the raw polarization images to restrict analysis to particle regions and suppress background artifacts.

## 2.5. Dataset Refinement by a feedback review loop

To establish a high-quality data foundation, the FRL is utilized to assess the dataset's consistency and identify outlier particles that could compromise model convergence. During this stage, a five-fold stratified cross-validation is performed. This evaluation is used to find images with errors that have a substantial negative impact on the classification training and evaluation accuracy.

Images are excluded only if they show substantial errors, specifically: (i) insufficient polarimetric signal area, (ii) air bubbles changing the polarization structure or (iii) artifacts from the manual focusing and preprocessing steps. These removals are only done on the training and validation data, not on the test set, to not artificially inflate the classification results for the used setup.

The FRL Dataset is split into five folds with preserved class distributions, so that a different 80/20 split of the data dedicated to training and validation is used in each iteration. This procedure is used to ensure that every particle image is evaluated in the validation set exactly once. Classification is carried out with a lightweight YOLO11n-based architecture (Ultralytics, 2023) initialized from pretrained weights. Hyperparameters are kept fixed across all folds (input size 640x640 pixels, batch size 32, 200 epochs, Adam optimizer). Only minor augmentations are used in the classification (see Supplementary Information D).

For each epoch, model predictions on the validation set are collected, and the per-image cross-entropy loss is computed as

$$L_i = -\log p\left(\mathrm{y_i} \mid \mathrm{x_i}\right) \qquad (Eq{:}\,2-1)$$

where $p(\mathrm{y_i} \mid \mathrm{x_i})$ is the predicted probability of the true class. This metric enables the identification of particle images that consistently produce high classification error across multiple folds. Images with exceptionally high average or maximum loss are flagged as problematic samples. These typically corresponded to the before mentioned errors.

The cross-validation procedure serves two important purposes: First, it functions as a dataset quality control measure, checking for systematic errors, biases, or mislabeled samples within the dataset. Second, it allows for particle-level error analysis by identifying which microplastic fragments generate the greatest confusion for the classifier. Crucially, this initial FRL is not intended to report final classification accuracy but rather to establish a reliable dataset foundation for subsequent training stages.

The main training path repeats the identical five-fold stratified cross-validation procedure executed in the FRL, but applies it based on the cleaned, refined dataset results. This process serves to evaluate the influence of error images on the classification model and validates the effectiveness of the dataset refinement. The data partitions and all hyperparameters remain strictly fixed. The validation set's purpose remained the monitoring of training progress via per-image cross-entropy loss calculation and the implementation of early stopping, which allows for the objective selection of the single best-performing model instance for each feature set (AOLP and DOLP) from the five folds. The test set remains completely untouched throughout this entire process.

The subsequent evaluation methodology focuses on a final analysis using entirely unknown data. Following the selection of the best-performing model from the main training path cross-validation, two final, critical analyses are conducted on the 15% test set. The first analysis is an unbiased performance estimation, where the selected model is evaluated only once on the test set. This provides an objective, final estimate of the classifier's generalization capability.

To determine the relative importance of distinct visual cues, specifically shape, structure, and texture, captured by the AOLP and DOLP feature sets, a systematic feature degradation study is performed. This analysis is executed by measuring the drop in classification accuracy relative to the baseline performance, which is established on the pristine original images (defined as shape + texture + context). Five distinct degradation scenarios are implemented to isolate the contribution of individual features. The structure & contour scenario involved replacing the particle's internal structure and texture with a uniform gray value, thereby isolating the contribution of the original, fine boundary shape. Conversely, the texture jitter test preserved the particle's overall shape and intensity histogram while destroying the local pixel arrangement by randomly shuffling pixel values within the particle's mask. To isolate the macro-shape, the uniform shape / hull scenario aggressively smoothed the particle's complex, fine boundary and filled the resulting shape uniformly, leaving only the approximate macro-shape as a feature. Finally, the boundaries of performance are established using the full noise scenario, where the entire image (particle and background) is corrupted with random noise to define a lower performance limit.

# 3. Results

The model's classification performance is evaluated across the integrated pipeline: the FRL establishes a baseline using five-fold stratified cross-validation on the original dataset, and the main training path assesses performance post-refinement. The key performance metrics for all four validation runs are consolidated in Table 2, while the full fold-by-fold results are provided in the Supplementary Information E: Tables E1–E4.

## 3.1. Feature Set Predictive Capacity

Compared to DOLP the AOLP feature set provides a stable signature. The model achieves a higher and more consistent classification rate, evidenced by a mean maximum top-1 accuracy of 0.85 ± 0.03 (Table 2). Generalization ability, assessed via the mean minimum validation loss, is stable at 0.43± 0.11 across the folds, indicating a reliable fit to the validation data with only small variance. In contrast, the DOLP feature set demonstrates lower predictive capacity and increased performance variability (Table 2). The mean maximum accuracy is significantly lower at 0.74 ± 0.06, representing a 0.11-point decrease compared to the AOLP set. Furthermore, the mean minimum validation loss is markedly higher (0.72 ± 0.11), suggesting poorer model generalization and greater sensitivity to data partitioning.

*Table 2: Consolidated Summary of five-fold Cross-Validation Performance across one feedback loop*

| **Loop** | **Feature Set** | **Mean Max Acc (Top-1)** | **Mean Min Val Loss** |
|---|---|---|---|
| Initial (FRL) | AOLP | 0.85 ± 0.03 | 0.43 ± 0.11 |
| | DOLP | 0.74 ± 0.06 | 0.72 ± 0.11 |
| Refined (Main) | AOLP | 0.89 ± 0.03 | 0.31 ± 0.09 |
| | DOLP | 0.81 ± 0.03 | 0.57 ± 0.10 |

The dataset refinement procedure resulted in the identification and removal of 30 problematic images (representing approximately 6% of the total dataset); representative examples of these outliers are shown in Figure 5.

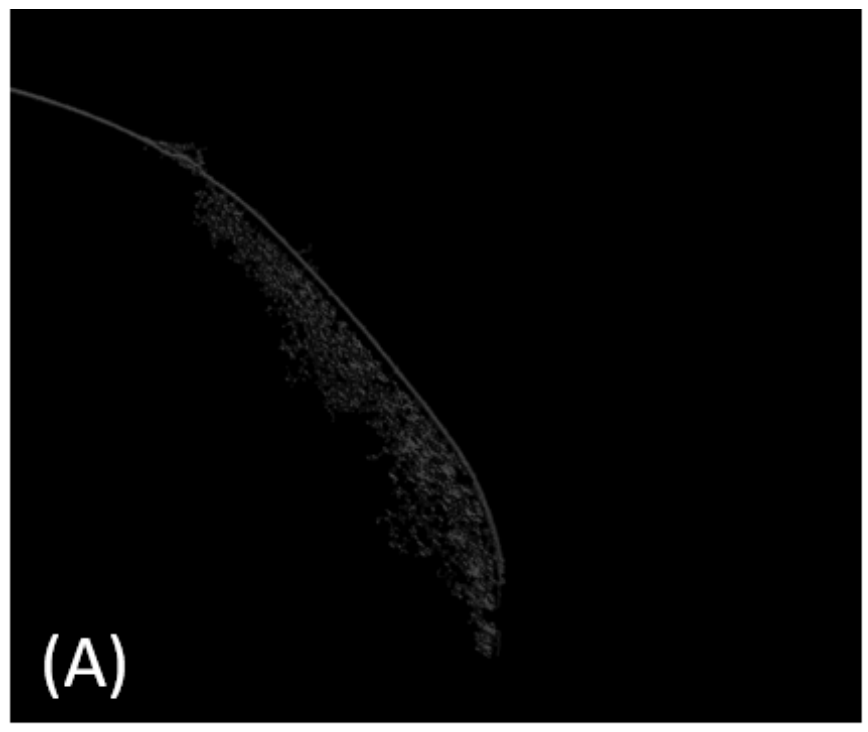

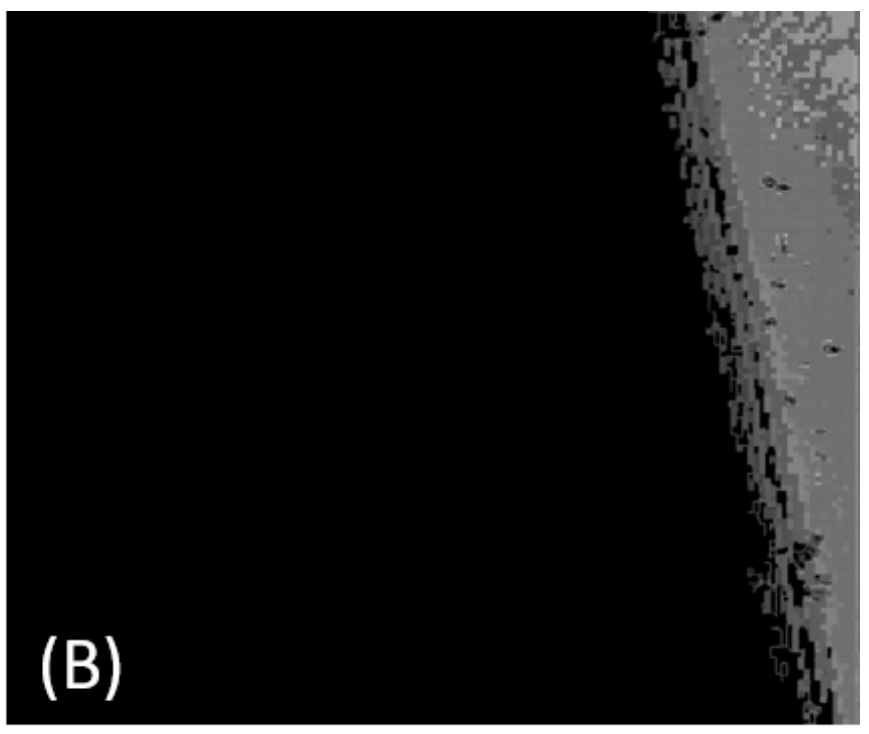

*Figure 5: Representative examples of data outliers identified during the FRL. (A) Particle rejection due to air bubble occlusion of the polarimetric signal. (B) Automated masking failure resulting in image artifacts.*

Based on consistently high validation loss, flagged images were reviewed and subsequently 30 images were removed from the training/validation image pool. It is important to note that while some of these flagged images were technically classified correctly, the disproportionately high loss indicated a state of high model uncertainty caused by optical artifacts rather than genuine material features. The dataset was re-evaluated via five-fold stratified cross-validation. The model performance improves significantly on the refined dataset. The mean maximum top-1 accuracy increases to 0.89 ± 0.03, demonstrating a gain in classification reliability. The DOLP feature set also exhibits stability improvement after data refinement, though its performance remained lower than AOLP. The mean maximum Top-1 accuracy increased to 0.81 ± 0.03.

## 3.2. Generalization on an unseen data (Test Set)

While the 0.89 accuracy serves as a metric for training consistency, the true predictive capability of the system is defined by its performance on the untouched 15% test set. On these entirely unseen samples, the AOLP model achieved a final classification accuracy of 0.80.

The delta between the refined training stability (0.89) and the test generalization (0.80) is expected in real-world aquatic sensing, as the test set contains the same "bubble noise" and focus variations that were removed from the training path. This demonstrates that while the model was trained on a clean foundation to learn polymer-specific textures, it remains robust when encountering uncleaned, raw environmental data.

This also confirms that the model did not overfit during the refinement process. The high degree of consistency between the validation and test accuracies proves that the "Human-in-the-Loop" cleaning did not bloat the results, but rather allowed the model to learn more stable features. The detailed classification error distributions for these top-performing models are visualized in the confusion matrices (see Figure 6).

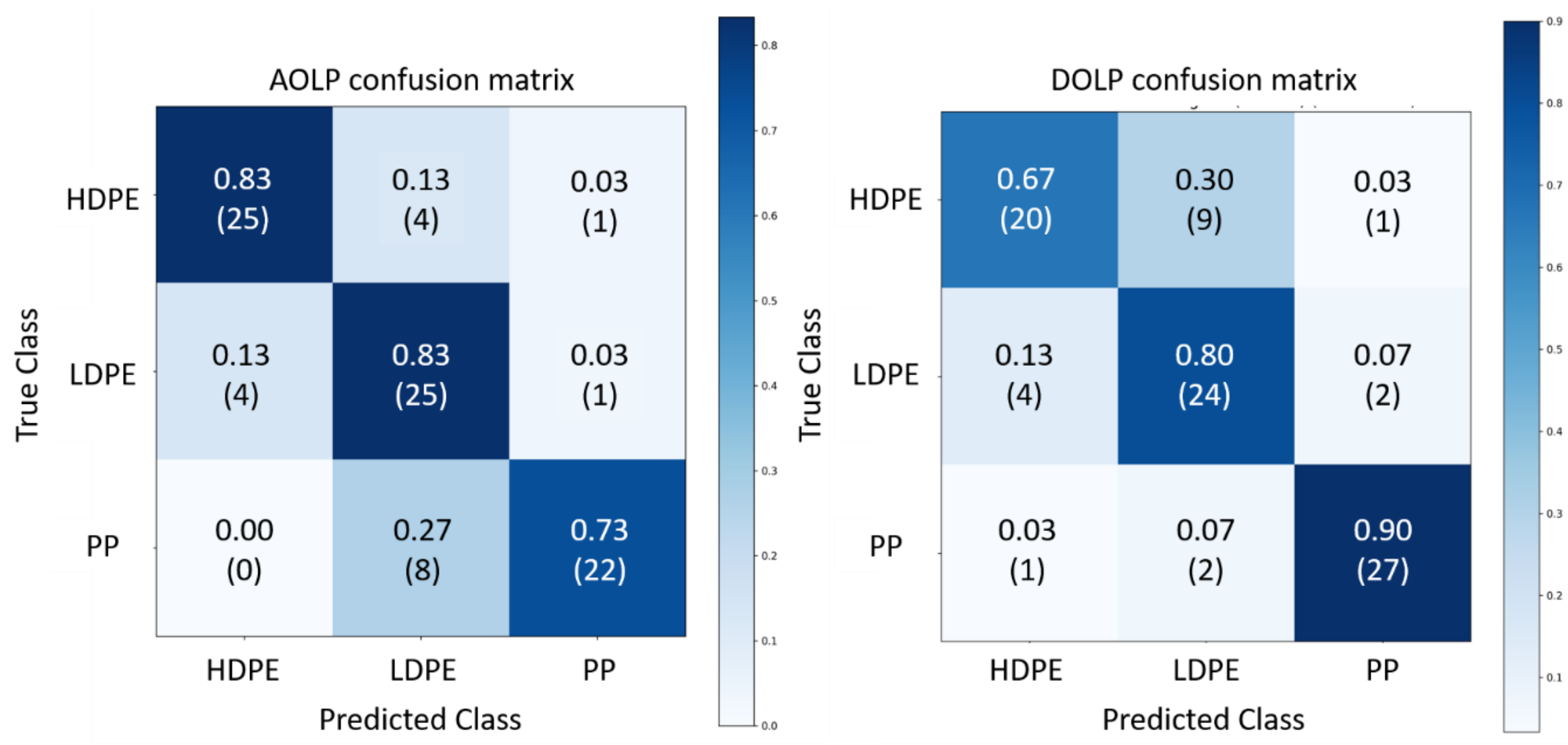

*Figure 6: Normalized confusion matrices for the classification on the best performing folds for AOLP and DOLP*

The AOLP model's primary source of error is the misclassification of PP samples as LDPE, which accounted for eight samples of the PP class errors. Additional confusion is observed between the two polyethylene variants: Four samples of HDPE are misclassified as LDPE, and four samples of LDPE are misclassified as HDPE. In contrast, the DOLP confusion matrix indicates a significantly higher error rate for the HDPE class, with nine samples being incorrectly predicted as LDPE. However, the DOLP model achieves the highest classification accuracy for the PP class, 0.90, compared to the 0.73 accuracy achieved by the AOLP model for the same material.

## 3.3. Feature Hierarchy Quantification and Signal Robustness

To identify the visual cues driving the classification, specifically the relative contributions of macro-shape (S), internal texture (T), and background context (C), a systematic feature degradation study was performed. This analysis quantifies the sensitivity of the CNN to the loss of specific polarimetric information and serves as a critical validation that the model is extracting material-specific signatures rather than merely memorizing particle silhouettes.

While research on general-purpose datasets like ImageNet has suggested that pre-trained CNNs naturally exhibit a bias toward internal texture rather than macro-contours (Geirhos et al., 2019), such general findings cannot be blindly applied to the specialized domain of polarimetric microplastic sensing. In our case, where the "textures" consist of complex, polarimetric reflections rather than standard visual patterns, it is essential to empirically demonstrate that these specific physical signals are indeed the primary drivers of classification.

The results, summarized in Table 3 and visually presented in Figure 7, reveal that the AOLP model relies primarily on the fine-grained structural properties of the polarization field. While the removal of internal texture in scenario 2 resulted in a minor performance decrease to 0.76, the corruption of local pixel arrangements in the Texture Jitter test (Scenario 3) caused a more significant drop to 0.70. The most dramatic collapse occurred when fine boundary details were smoothed into a uniform hull in Scenario 4, where accuracy fell to 0.38. This suggests that for the AOLP signal, the discriminative material information is stored within the high-frequency spatial transitions and the precise, structured edges of the polarized reflection.

*Table 3: Feature Hierarchy Quantification via Feature Degradation Tests for the AOLP Signal.*

| Scenario | Accuracy | F1-Score (Macro) | Avg. Confidence |
|---|---|---|---|
| 1. Original (S + T + C) | 0.80 | 0.80 | 0.91 |
| 2. Structure & Contour (S + C) | 0.76 | 0.75 | 0.83 |
| 3. Texture Jitter (S + $T_{corrupted}$ + C) | 0.70 | 0.69 | 0.88 |
| 4. Uniform Shape ($S_{smoothed}$) | 0.38 | 0.36 | 0.77 |
| 5. Full Noise | 0.36 | 0.21 | 0.98 |

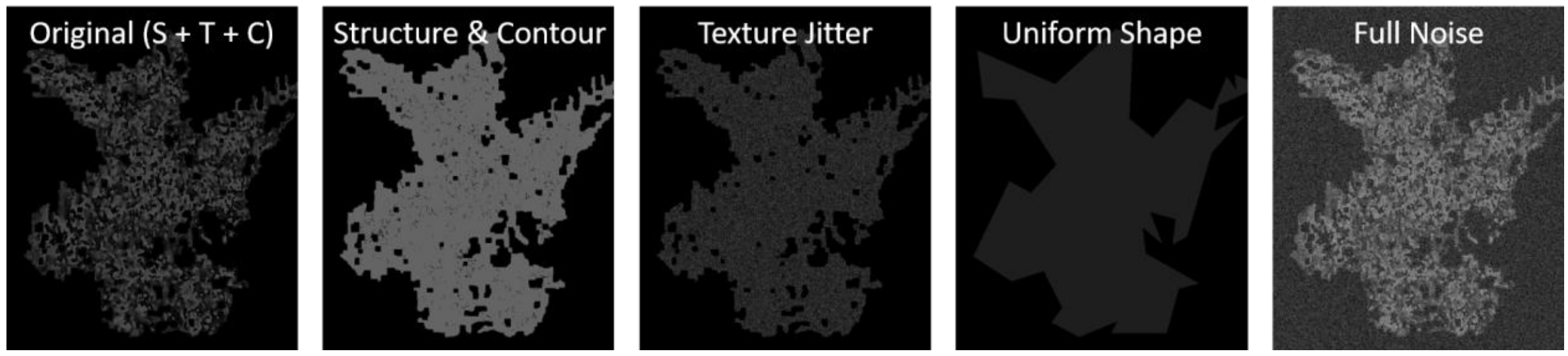

*Figure 7: Feature-Degraded Image Scenarios for AOLP Classifier Robustness Assessment. It shows on a sample PP particle image the original together with the four degradation scenarios used to quantify the feature hierarchy (Shape, Texture, and Context) and test the robustness of the classifier.*

As shown in Table 4, the DOLP model demonstrates an even more critical dependence on internal structural integrity (Figure 7). The transition from the original image to a uniform internal structure in scenario 2 caused a massive accuracy collapse from 0.79 to 0.61. Unlike the AOLP model, which maintained moderate performance based on shape alone, the DOLP model's predictive power appears almost entirely based on internal textural intensity. This reliance was further confirmed by the Texture Jitter test, which reduced performance to 0.51, indicating that the spatial correlation of the degree of polarization is the primary carrier of material information for this feature set.

*Table 4: Feature Hierarchy Quantification via Feature Degradation Tests for the DOLP Signal.*

| Scenario | Accuracy | F1-Score (Macro) | Avg. Confidence |
|---|---|---|---|
| 1. Original (S + T + C) | 0.79 | 0.79 | 0.88 |
| 2. Structure & Contour (S + C) | 0.61 | 0.61 | 0.78 |
| 3. Texture Jitter (S + $T_{corrupted}$ + C) | 0.51 | 0.48 | 0.84 |
| 4. Uniform Shape ($S_{smoothed}$) | 0.39 | 0.35 | 0.78 |
| 5. Full Noise | 0.33 | 0.17 | 0.97 |

The baseline accuracy for the uncorrupted DOLP image (Test 1: S+T+C) is 0.79. The feature degraded images are shown in Figure 8.

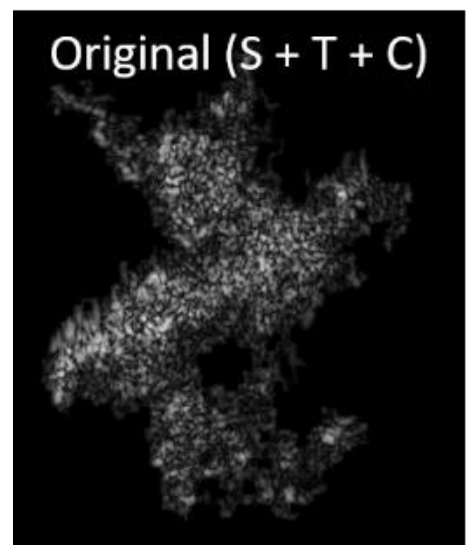

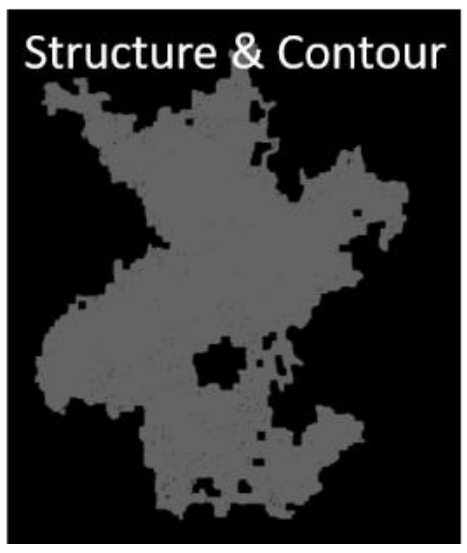

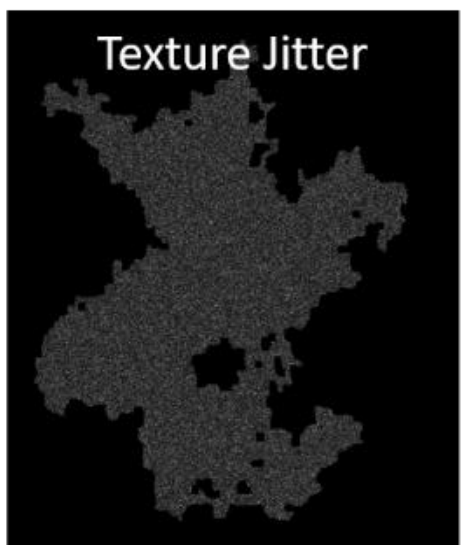

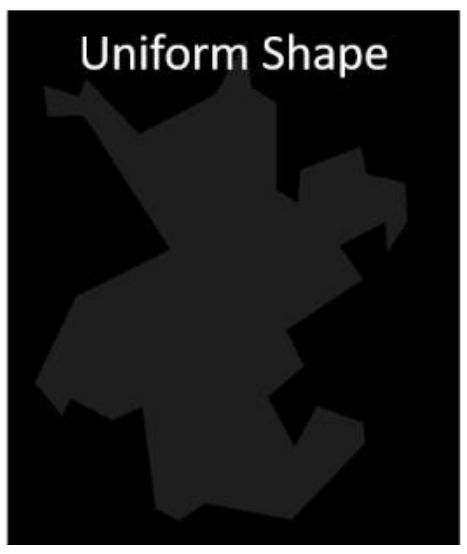

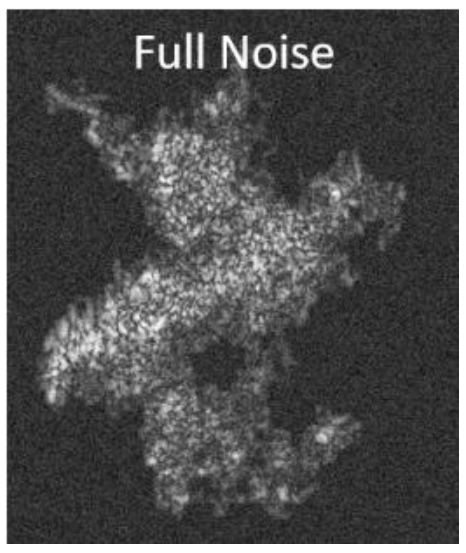

*Figure 8: Feature-Degraded Image Scenarios for DOLP Classifier Robustness Assessment. Shows on a sample PP particle image the original together with the four degradation scenarios used to quantify the feature hierarchy (Shape, Texture, and Context) and test the robustness of the classifier.*

These results of the feature hierarchy provide a robust defense of the proposed methodology. By demonstrating that classification performance collapses toward the noise floor (0.33–0.36) when micro-textures are removed, we prove that the system is successfully extracting optical signatures that are invisible to standard, non-polarimetric imaging. While standard visual classification is limited by morphological ambiguity, our results confirm that the CNN prioritizes internal polarimetric textures: the DOLP signal acts as a high-sensitivity carrier for internal material structure, while AOLP provides a stable, intensity-focused structural map.

## 3.4. Signal combination by late stage fusion

Given that the AOLP and DOLP signals exhibit complementary strengths, with AOLP excelling in polyethylene differentiation and DOLP demonstrating superior precision for polypropylene, a late fusion architecture was implemented to combine these material-specific descriptors.

The late stage fusion model architecture is designed to integrate high-dimensional features from dual polarimetric channels (AOLP and DOLP). The Multi-Layer Perceptron (MLP) fusion head acts as a meta-classifier that synthesizes the descriptors extracted from the frozen YOLO backbones. The specific layer-by-layer configuration, including neuron counts and activation functions, is detailed in supplementary information F. The fusion is achieved by concatenating the feature vectors from the final convolutional layers of both the AOLP and DOLP branches, allowing the MLP to learn an optimized decision boundary based on the combined inputs.

As illustrated in the late stage fusion confusion matrix (see Figure 9), this ensemble approach yields a significantly more balanced performance across all polymer classes.

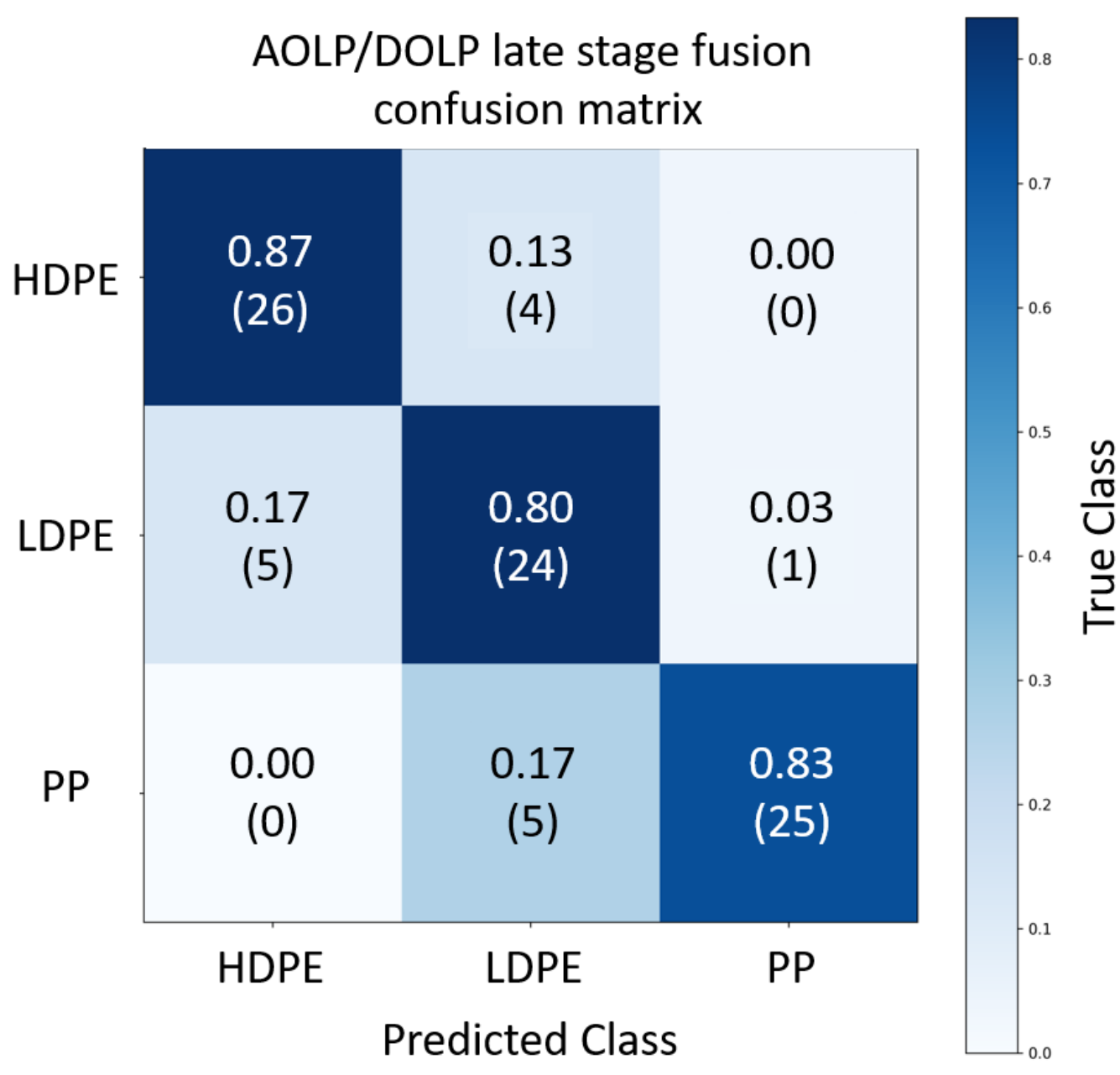

*Figure 9: Normalized confusion matrix for the late stage fusion classification*

The model effectively mitigates the individual weaknesses of each separate signal, achieving a stable accuracy of 87% for HDPE, 80% for LDPE, and 83% for PP. This architecture preserves the high sensitivity to polypropylene found in the DOLP model while simultaneously leveraging the structural clarity for polyethylene variants provided by the AOLP signal. While the computational load is increased due to the simultaneous processing of dual feature maps, the resulting gain in classification reliability is measurable. This late stage integration ensures that the system is not reliant on a single optical descriptor, making it far more robust for real-world environmental monitoring where various polymer types are encountered in mixed populations.

# 4. Discussion

The interpretation of the classification results must first be considered in the context of the underlying imaging geometry and polarimetric design. The choice of a 120° backscattering reflection configuration represents a departure from transmission-based approaches, enabling the analysis of opaque and irregularly shaped microplastic particles directly in aqueous environments. Previous studies have shown that backscattered light at oblique angles is particularly sensitive to variations in refractive index and surface morphology (Boss and Pegau, 2001; Koestner et al., 2024; Liu et al., 2021), making it well suited for resolving material-dependent scattering signatures under realistic conditions.

The use of snapshot DoFP polarimetric imaging further constrains the nature of the extracted features. While this approach enables high-throughput acquisition at the hardware framerate, it also imposes limitations in spatial resolution and polarization sampling compared to rotating-analyzer systems (Mihoubi et al., 2018; Qiu et al., 2021). Consequently, the classification framework relies on spatially distributed polarization patterns rather than high-precision point measurements. Within this design space, the selection of AOLP and DOLP as primary descriptors reflects a balance between robustness and sensitivity: AOLP provides a stable, intensity-independent representation of scattering orientation, while DOLP captures depolarization effects linked to surface roughness and microstructure (Collett, 2005; Kupinski et al., 2019).

Against this physical and instrumental background, the observed classification performance can be interpreted as a direct consequence of how these polarimetric signals encode material-specific information.

The classification performance across all evaluation stages reveals a clear hierarchy in the predictive capacity and robustness of the extracted polarimetric features. In particular, the superior performance of the AOLP-based model, which consistently outperforms DOLP by approximately 8 percentage points, indicates that AOLP encodes a more stable and transferable material descriptor. This observation is consistent with the physical formulation $(Eq: 1-1)$ and $(Eq: 1-2)$ of polarization parameters, where AOLP is independent of total intensity ($S_0$) and therefore inherently less sensitive to fluctuations in illumination or scattering conditions. In contrast, DOLP directly depends on intensity ratios and is thus more susceptible to environmental perturbations such as turbidity or local intensity variations (Collett, 2005).

The effectiveness of the FRL further highlights the importance of data quality in polarimetric learning frameworks. By selectively removing samples affected by air bubble interference or preprocessing artifacts, the model was prevented from overfitting to non-physical signal distortions. Similar challenges in microplastic analysis, particularly contamination, sample handling artifacts, and optical ambiguities, have been widely reported in the literature (Brander et al., 2020). The improvement in validation stability after FRL therefore supports the notion that careful dataset curation is essential when training data-driven models on physically complex optical signals. Although the resulting dataset is relatively compact, the high degree of feature consistency achieved through the FRL proved sufficient for this fundamental evaluation of the polarimetric framework. However, the relatively small dataset size remains a limitation, and future work should prioritize larger and more diverse datasets to improve generalization across environmental conditions. Furthermore, all measurements were conducted under controlled laboratory conditions with purified water and isolated particles. In natural aquatic environments, additional factors such as biofouling, organic matter, and heterogeneous suspended solids may alter the polarimetric response and introduce signal ambiguity not captured in the present study.

A key finding of this study is that classification performance is primarily driven by internal polarimetric texture of the images, rather than external morphology. The feature degradation experiments demonstrate that removing internal structure leads to a collapse in accuracy toward near-random performance, confirming that the CNN relies on sub-resolution scattering patterns linked to material microstructure. Importantly, in the context of microplastic classification, this behavior is advantageous: unlike visual appearance, which can vary significantly due to environmental weathering (Jahnke et al., 2017), internal scattering properties are more directly tied to polymer composition and structure.

Despite this shared reliance on texture, the AOLP and DOLP features exhibit distinct functional characteristics. The DOLP-based model shows a significantly higher sensitivity to structural perturbations, indicating a stronger dependence on fine-scale surface roughness and local scattering variations. This is consistent with the physical interpretation of DOLP as a measure of polarization purity, which is strongly influenced by depolarization effects arising from microscopic surface irregularities. As a result, DOLP provides higher material sensitivity but lower robustness. In contrast, AOLP captures more spatially coherent polarization orientations, leading to greater stability but slightly reduced sensitivity to subtle surface differences. These differences are also reflected in the class-specific performance. While AOLP achieves better separation between chemically similar polyethylene variants (HDPE and LDPE), DOLP demonstrates superior discrimination of polypropylene. This complementary behavior supports the interpretation that the two signals encode non-redundant physical information, as also suggested in previous studies on polarimetric imaging for material classification (Hu et al., 2016; Kupinski et al., 2019).

From a methodological perspective, the snapshot-based acquisition used in this study represents a significant advantage over conventional spectroscopic approaches. Techniques such as Raman spectroscopy, while highly specific, are fundamentally limited by low scattering cross-sections and long acquisition times (Kniggendorf et al., 2019; Schymanski et al., 2021). In contrast, the use of reflected polarized light enables rapid image acquisition at the hardware framerate, making the approach inherently suitable for high-throughput and real-time monitoring applications. This capability is particularly relevant for dynamic environments, including marine systems or medical infusion monitoring, where transient contamination events may otherwise go undetected.

The integration of AOLP and DOLP through a late fusion architecture further demonstrates the benefit of combining complementary polarimetric features. By merging structurally robust AOLP descriptors with the surface-sensitive DOLP signal, the model is able to balance environmental robustness with material specificity. The resulting improvement in classification performance confirms that multimodal polarimetric analysis can overcome the limitations of single-feature approaches. This finding position reflection-based polarimetry as a promising intermediate solution between highly specific but slow laboratory techniques and rapid but non-specific bulk optical sensors.

Overall, the results demonstrate that spatially resolved polarimetric imaging, when combined with deep learning, enables the extraction of material-specific features that are inaccessible to conventional intensity-based imaging. This provides a strong foundation for the development of in-situ microplastic monitoring systems capable of operating in complex and dynamic aquatic environments.

# 5. Summary & Outlook.

This study presents a reflection-based 120° backscattering polarimetry system with late-fusion CNN classification, achieving 83% accuracy for identifying opaque PP/HDPE/LDPE microplastic fragments (50-300 µm) directly in water without filtration or drying. The pipeline integrates Division-of-Focal-Plane polarization sensors, Fourier-domain demosaicing for Stokes parameter reconstruction, and a FRL that systematically removes 6% high-loss outliers to yield 9% val loss improvement compared to baseline training, demonstrating that reflection-based polarimetric imaging provides a robust solution for in-situ microplastic classification.

Central mechanistic insight from feature hierarchy analysis reveals classification is driven primarily by internal microstructural textures of the images rather than macro-morphology, making the approach shape-agnostic. The FRL proved critical for stabilizing performance by mitigating data heterogeneity in this relatively small but highly consistent dataset. While AOLP emerged as the more resilient feature against environmental noise due to $S_0$-independence, late fusion of complementary AOLP and DOLP signals remains essential for resolving chemically similar polymers like PP and PE.

To transition this laboratory-validated framework toward autonomous field applications, several key research areas must be addressed. A primary objective is the expansion of the object library to include a wider diversity of materials and scattering geometries, particularly microplastic fibers, but account for the complexity of real-world aquatic matrices. In this context, the generation of larger and more heterogeneous datasets will be crucial to further enhance model generalization. A significant opportunity for scaling these datasets lies in the automation of the image acquisition process. Leveraged by the inherent advantages of the snapshot methodology, which captures complete polarimetric information in a single frame, the system enables high-speed, automated data generation. Implementing an automated pipeline for rapid image capturing and labeling will not only increase the volume of training data but also refine the statistical robustness of future evaluations.

Furthermore, since environmental particles are rarely pristine, assessing the resilience of polarimetric textures against surface modifications, such as biofouling or oxidation, is the next logical step. Here, the integration of object detection and segmentation frameworks could improve classification performance by isolating fouled versus clean particle regions. Employing models beyond conventional YOLO, such as Vision Transformers, alternative CNN architectures, or hybrid approaches, would enable more flexible and accurate extraction of polarimetric features across heterogeneous samples.

Additionally, algorithmic refinements offer significant potential, including transitioning the FRL into a fully automated outlier detection system or utilizing multi-spectral polarimetry for richer material information. Ultimately, validating the methodology on environmental samples is vital to ensure that the learned features remain stable for naturally weathered plastics, providing a scalable tool for the protection of global aquatic ecosystems.

# Acknowledgments

## Declaration of generative AI and AI-assisted technologies in the manuscript preparation process

During the preparation of this work, the authors used AI tools to improve clarity and fluency in the writing process. After using these tools, the authors reviewed and edited the content as needed and take full responsibility for the content of the published article.

## Data availability

The microplastics dataset used to support the findings of this study, including the raw intensity images and derived Stokes/AOLP/DOLP images, is openly available in the Zenodo repository and can be accessed via the permanent Digital Object Identifier (DOI):
https://doi.org/10.5281/zenodo.18985055

# Supplementary Information: Classification of Microplastic Particles in Water using Polarized Light Scattering and Machine Learning Methods

## A. Laser stability and beam divergence

Prior to the measurements, the output power of the 532 nm diode laser (MD532-1-5(20×80), PicoTronics) was assessed for temporal stability. Measurements were performed using a calibrated optical power meter (Model: [LaserChecker™/COHERENT]), both directly at the laser exit and at the sample position. After a 15-minute warm-up phase, the laser output stabilized at 1 mW (±SD 0.01 mW). Stability was monitored over a 2-hour period at 10-minute intervals under ambient laboratory conditions (22-27°C). The observed variance remained below 1%, with no significant fluctuations detected. The diode laser is specified with a beam divergence of 0.5 mrad; in the setup, the beam was focused through a collimating lens to ensure uniform illumination of the particle field of view.

## B. Polarization de-mosaicing via Fourier-Domain Interpolation

The Division-of-Focal-Plane (DoFP) sensor arrangement results in four spatially sub-sampled intensity images, $I_\theta(x,y)$, where $\theta \in \{0°,45°,90°,135°\}$. Each image Iθ has a spatial resolution of H/2×W/2 and contains only the directly measured pixels, with no zero-padding applied (as confirmed by the sensor configuration). The objective of demosaicing is to reconstruct the four full-resolution (H×W) intensity images, Iθ′.

Our method employs Fourier-domain zero-padding (FDZP), which achieves sinc-interpolation, the theoretically ideal reconstruction function for band-limited signals (Eskreis-Winkler et al., 2017). The process for each sub-sampled polarization channel Iθ is as follows:

1. Fourier Transform: The image is transformed into the frequency domain:

$$\mathrm{F\theta(u,v) = F\{I\theta(x,y)\}} \qquad (Eq\!: A-1)$$

2. Zero-Padding (up-sampling): The low-resolution frequency spectrum Fθ is embedded into a higher-resolution grid (size H×W) by padding the high-frequency quadrants with zeros. This process retains the integrity of the sampled low-frequency information while defining the high-frequency components as zero, consistent with the ideal sinc function reconstruction:

$$\mathrm{F\theta'(u,v) = Zero - Pad\{F\theta(u,v)\}} \qquad (Eq\!: A-2)$$

3. Inverse Fourier transform: The full-resolution image is obtained by the inverse transform:

$$\mathrm{I\theta'(x,y) = F - 1\{F\theta'(u,v)\}} \qquad (Eq\!: A-3)$$

This FDZP approach effectively performs a nearest-neighbor interpolation in the frequency domain, resulting in sinc interpolation in the spatial domain. A known consequence of this theoretically ideal interpolation is the introduction of Gibbs ringing artifacts at sharp material or polarization boundaries in the final images, a factor that is generally less critical in the analysis of water bodies compared to high-detail terrestrial scenes (Mihoubi et al., 2018; Qiu et al., 2021). The resulting $I\theta'$ are then used to calculate the Stokes parameters.

## C. Image preprocessing and particle segmentation

Raw polarization-resolved intensity images were exported directly from the Lucid Vision SDK. All internal auto-functions (white balance, auto-exposure, auto-gain) were disabled.

The segmentation workflow proceeded as follows:

1. Preprocessing:
    a. Conversion to grayscale.
    b. Non-local means filtering to reduce high-frequency noise while preserving edge information.
    c. Local contrast enhancement using CLAHE (clip limit = [2.0]).
2. Edge detection:
    a. Canny edge detector with low/high thresholds = 30/100.
3. Morphological refinement:
    a. Closing to connect fragmented edges.
    b. Dilation to reinforce particle boundaries.
    c. A second closing step to ensure filled masks.
4. Connected-component analysis:
    a. Largest connected component retained.
    b. Resulting binary mask applied to the original grayscale image.

This pipeline produced consistent particle-only masks with background suppression across all samples.

## D. Classification details

For classification, DOLP and AOLP images were treated as separate input channels. Training was conducted using the YOLO11-nano cls architecture (Ultralytics, 2023).

- Dataset split: 85% train and validation, 15% test. Particles were uniquely assigned to one split to avoid leakage.
- Labels: Ground-truth labels derived from source-verified single-material samples embedded in distilled water.
- Training setup:
    - Input resolution: 640×640 px.
    - Epochs: 200.
    - Early stopping: 50
    - Batch size: 32.
    - Optimizer: Adam.
    - Initialization: ImageNet pretraining.
    - Data augmentation: Erasing and minor intensity changes
- Evaluation: Performance metrics included top-1 accuracy, class-specific confidence scores, and confusion matrices.

- Runtime environment: Training performed on a local workstation with PyTorch and the Ultralytics API. Logs, models, and predictions were archived for reproducibility.

## E. Results of five-fold stratified cross-validation on original dataset:

Table E.1 AOLP FRL five-fold stratified cross-validation results

| Fold | Epoch (Min Val Loss) | Min Val Loss | Train Loss (at Min Val Loss) | Acc. (at Min Val Loss) | Epoch (Max Acc) | Max Acc (Top1) |
|---|---|---|---|---|---|---|
| 0 | 60 | 0.28 | 0.23 | 0.85 | 22 | 0.90 |
| 1 | 22 | 0.54 | 0.41 | 0.79 | 32 | 0.83 |
| 2 | 38 | 0.50 | 0.42 | 0.78 | 70 | 0.84 |
| 3 | 22 | 0.44 | 0.44 | 0.80 | 48 | 0.84 |
| 4 | 24 | 0.41 | 0.41 | 0.76 | 77 | 0.83 |
| Mean ± Std Dev | N/A | 0.43 ± 0.11 | 0.38 ± 0.08 | 0.81 ± 0.03 | N/A | 0.85 ± 0.03 |

Table E.2 DOLP FRL five-fold stratified cross-validation results

| Fold | Epoch (Min Val Loss) | Min Val Loss | Train Loss (at Min Val Loss) | Acc. (at Min Val Loss) | Epoch (Max Acc) | Max Acc (Top1) |
|---|---|---|---|---|---|---|
| 0 | 27 | 0.75 | 0.67 | 0.66 | 71 | 0.69 |
| 1 | 107 | 0.89 | 0.23 | 0.65 | 69 | 0.67 |
| 2 | 21 | 0.65 | 0.73 | 0.73 | 107 | 0.80 |
| 3 | 30 | 0.61 | 0.67 | 0.75 | 29 | 0.79 |
| 4 | 16 | 0.71 | 0.74 | 0.67 | 67 | 0.77 |
| Mean ± Std Dev | N/A | 0.72 ± 0.11 | 0.61± 0.22 | 0.69 ± 0.04 | N/A | 0.74 ± 0.06 |

Cleaned dataset:

Table E.3 AOLP Main five-fold stratified cross-validation results

| Fold | Epoch (Min Val Loss) | Min Val Loss | Train Loss (at Min Val Loss) | Acc. (at Min Val Loss) | Epoch (Max Acc) | Max Acc (Top1) |
|---|---|---|---|---|---|---|
| 0 | 93 | 0.24 | 0.31 | 0.91 | 92 | 0.92 |
| 1 | 81 | 0.42 | 0.20 | 0.85 | 81 | 0.85 |
| 2 | 40 | 0.33 | 0.41 | 0.85 | 17 | 0.88 |
| 3 | 87 | 0.35 | 0.19 | 0.88 | 71 | 0.88 |
| 4 | 36 | 0.19 | 0.37 | 0.88 | 33 | 0.92 |
| Mean ± Std Dev | N/A | 0.31±0.09 | 0.29±0.09 | 0.87±0.02 | N/A | 0.89±0.03 |

Table E.4 DOLP Main five-fold stratified cross-validation results

| Fold | Epoch (Min Val Loss) | Min Val Loss | Train Loss (at Min Val Loss) | Acc. (at Min Val Loss) | Epoch (Max Acc) | Max Acc (Top1) |
|---|---|---|---|---|---|---|
| 0 | 23 | 0.61 | 0.71 | 0.72 | 25 | 0.78 |
| 1 | 123 | 0.46 | 0.34 | 0.82 | 83 | 0.85 |

| 2 | 74 | 0.66 | 0.33 | 0.74 | 64 | 0.78 |
|---|---|---|---|---|---|---|
| 3 | 61 | 0.65 | 0.41 | 0.75 | 50 | 0.79 |
| 4 | 126 | 0.49 | 0.37 | 0.81 | 128 | 0.83 |
| Mean ± Std Dev | N/A | 0.57±0.10 | 0.43±0.16 | 0.77±0.04 | N/A | 0.81±0.03 |

# F. Detailed MLP Fusion Head Configuration

The MLP fusion head was designed to process concatenated features from the AOLP and DOLP branches (2 * $N_{Features}$). The sequential structure of the model is as follows:

- **Feature Integration:** Input features from both polarimetric backbones are concatenated into a single feature vector.
- **First Hidden Layer:** A linear transformation to 512 neurons, followed immediately by Batch Normalization to stabilize internal covariate shift and ReLU activation to introduce non-linearity.
- **Regularization:** A Dropout layer with a probability of 0.3 is applied to prevent overfitting by randomly zeroing elements during training.
- **Second Hidden Layer:** A linear transformation to 256 neurons followed by ReLU activation.
- **Output Layer:** A final linear transformation to 3 neurons, corresponding to the polymer classes (PP, HDPE, and LDPE). The final classification is determined by a Softmax function.

## Training and Implementation Parameters

The classification framework was implemented in PyTorch and executed on an NVIDIA GPU using CUDA-acceleration. The following settings were used for the training protocol:

- **Optimizer:** Adam (Adaptive Moment Estimation)
- **Learning Rate:** $10^{-4}$
- **Batch Size:** 32
- **Epochs:** 50
- **Criterion:** Cross-Entropy Loss
- **Validation Strategy:** Model checkpoints were saved based on the highest validation accuracy achieved during the 50-epoch cycle.
- **Evaluation Protocol:** Final performance metrics, including accuracy and confusion matrices, were generated by evaluating the best-performing model checkpoint once on the strictly held-out test set.